**Pilot Testing an Artificial Intelligence Algorithm That Selects Homeless Youth Peer Leaders Who Promote HIV Testing**

Eric Rice, PhD [a]

Robin Petering, MSW [a]

Jaih Craddock, MSW [a]

Amanda Yoshioka-Maxwell, MSW [a]

Amulya Yadav, MS [b]

Milind Tambe, Phd [b]

Unpublished manuscript: August 18, 2016

[a] School of Social Work, University of Southern California

[b] Viterbi School of Engineering, University of Southern California

Acknowledgements: Funding for this study was provided by the USC School of Social Work. The authors would like to thank the staff at Safe Place for Youth and the young people who participated in the project.

**Abstract**

*Objective.* To pilot test an artificial intelligence (AI) algorithm that selects peer change agents (PCA) to disseminate HIV testing messaging in a population of homeless youth.

*Methods.* We recruited and assessed 62 youth at baseline, 1 month ($n$ = 48), and 3 months ($n$ = 38). A Facebook app collected preliminary social network data. Eleven PCAs selected by AI attended a 1-day training and 7 weekly booster sessions. Mixed-effects models with random effects were used to assess change over time.

*Results.* Significant change over time was observed in past 6-month HIV testing (57.9%, 82.4%, 76.3%; $p$ < .05) but not condom use (63.9%, 65.7%, 65.8%). Most youth reported speaking to a PCA about HIV prevention (72.0% at 1 month, 61.5% at 3 months).

*Conclusions.* AI is a promising avenue for implementing PCA models for homeless youth. Increasing rates of regular HIV testing is critical to HIV prevention and linking homeless youth to treatment.

**Implications and Contribution**

Homeless youth are in great need of linkage to HIV testing and treatment. Artificial intelligence can be used to augment intervention delivery of peer-led dissemination models.  A pilot test with a pre-test post-test design resulted in a nearly 20% increase in the number of youth reporting recent HIV testing.

Despite the need for HIV prevention for homeless youth (HY), few evidence-based interventions exist for HY.[1] Given the important role peers play in the HIV risk and protective behaviors of HY,[2,3] it has been suggested that peer change agent (PCA) models for HIV prevention be developed for HY.[1-3]

PCA models have been effective for the prevention of HIV in many contexts,[4] but there have been some notable failures[5] that may be due to how the PCAs were selected to participate in the intervention.[6-8] Change agents can often be as important that the messages they convey. Rarely have network methods that select PCAs based on structural position been attempted.[6-8]

Selecting PCAs based on structural position requires: (a) the ability to "map" the network space of the target population and (b) a viable structural solution. Prior methods of collecting whole networks of homeless youth accessing drop-in centers required resources prohibitive to future community-based implementation.[4] Thus, an integrative Facebook app was developed to collect this information. Computer scientist partners developed an artificial intelligence (AI) algorithm to select PCAs that outperforms other structural network PCA selection rules suggested by Schneider,[6] such as degree or betweenness centrality.[7-8]

This paper presents results of a pilot study of an AI-enhanced PCA prevention program for homeless youth. In accordance with the field's push to engage underserved populations in the HIV continuum of care, PCA training and peer messaging focused on increasing regular HIV testing (every 3 to 6 months).

## METHODS

### Recruitment

All study procedures were approved by the [blinded for review] institutional review board. Sixty-two youth (aged 16–24) seeking drop-in homelessness support services (e.g., food, clothing, case management, mobile HIV testing site) in Los Angeles were recruited into the study. All youth receiving services were eligible to participate and were informed of the study as they entered the drop-in center. Participants were required to have a Facebook profile, although there were no requirements regarding how often they use it and if a participant did not have a Facebook account they could create one ($n$ = 5).

**Assessments**

Participants completed a computer-based self-administered survey at baseline ($n$ = 62), 1 month ($n$ = 48, 77.4%), and 3 months ($n$ = 38, 61.3%) and received $20, $25, and $30 for each respective assessment.

**Network Data**

A Facebook app collected network data regarding which participants were connected to one another, i.e., friends. No information about individuals who were not study participants was collected by the app, which did not appear on their Facebook profiles in any way. These data were augmented by field observations collected by the research team during the 2 weeks of recruitment, based on which participants regularly interacted with one another.

**AI-Based Peer Selection**

Papers detailing the development and computational experiments of the AI algorithm exist.[7,8] The algorithm selects a set of the best likely PCAs who can maximize influence in the network at a given point in time. Only four PCA could be trained at once and thus PCAs were enrolled in three subsequent rounds. The Facebook and field observation are incomplete (e.g., not all Facebook friends still communicate, not all persons in the network connect on Facebook with all their social ties, many relationships cannot be observed by research staff). The AI algorithm is the first to explicitly model this uncertainty of information.[7,8] The first step of the algorithm is to select a given network among thousands of possible errors in network data collection. It then selects the best PCA set for that particular network. It then selects among the resulting millions of possible combinations of PCAs and networks to arrive at a best solution of HY to be recruited by staff to be trained as PCAs. In part, the algorithm outperforms other structural solutions because AI remembers who was picked last time, considers who they could possibly reach, and then picks the next set of actors to maximize coverage in areas of the network where youth are unlikely to have been influenced by the previously trained PCA.

**Intervention Training and Delivery**

Training lasted approximately 6 hours and was facilitated by three researchers. Training was interactive and broken into six hour-long modules on sexual health and condom use; HIV, hepatitis C, and sexually transmitted infection facts; communication skills; outreach techniques; and leadership skills. PCA were asked to focus their outreach efforts on other youth in the private Facebook group who were all study participants and to promote regular HIV testing. Seven weekly booster sessions allowed PCA to discuss successes and potential outreach barriers and reinforced training content. PCAs received $60 for the training and $20 for each booster session.

**Statistical Analysis**

Longitudinal analysis was conducted using mixed-effects models, with random effects, using the GLIMMIX procedure in SAS 9.4.[9]

**RESULTS**

Nine (12.68%) youth approached declined to participate in the study. Reasons primarily involved unwillingness to provide Facebook information. The algorithm selected eight youth as PCAs during each of the 3 following weeks. Of the 24 youth selected, 16 were successfully contacted and 11 participated. Only one declined to participate.

Most study participants reported having a conversation about HIV prevention with a PCA. There was a significant increase over time in HIV testing but not condom use.

**Conclusions**

The follow-up rate is similar to that of other recent interventions involving homeless youth in this age range.[10] Given the transience of homeless youth, we were very successful in engaging youth as PCAs (only one declined). Despite the small sample size, a significant change in recent HIV testing behavior was observed. Condom use was not the primary focus of the PCA messaging, which may account for the lack of change in condom use.

The limitations of this study include the lack of comparison group, small sample size, modest loss to follow-up, and limited generalizability given the sample was recruited from a single drop-in center. Future research should include two control groups, a standard PCA selection protocol (volunteers or staff recommendations) and a control group without intervention, to assess the effect of repeated surveying on recent HIV testing.

TABLE 1—Background Characteristics and Outcomes among Homeless Youth (N = 62), Los Angeles, CA, 2016

|                                          | %          |
| ---------------------------------------- | ---------- |
| **Background characteristics**           |            |
| Gender                                   |            |
| Male                                     | 75.8       |
| Female                                   | 22.6       |
| Transgender                              | 1.6        |
| Race and ethnicity                       |            |
| Asian American                           | 5.0        |
| African American                         | 20.0       |
| Native Hawaiian or Pacific Islander      | 3.3        |
| White                                    | 41.7       |
| Latino                                   | 13.3       |
| Mixed                                    | 16.7       |
| Sexual orientation                       |            |
| Homosexual (gay or lesbian)              | 3.3        |
| Bisexual                                 | 16.4       |
| Heterosexual (straight)                  | 78.7       |
| Questioning or unsure                    | 1.6        |
| Age[a]                                   | 21.7 (2.3) |
| **Outcomes**[b]                          |            |
| HIV test in past 6 months*               |            |

Baseline                                                    57.9

  1-month follow-up                              82.4

  3-month follow-up                              76.3

Unprotected sex

  Baseline                                       63.9

  1-month follow-up                              65.7

  3-month follow-up                              65.8

Target youth receiving messages in past month

  1-month follow-up                              72.0

  3-month follow-up                              61.5

---

[a]Figures represent mean and standard deviation.

[b]Percentages reflect data from 38 youth with complete follow-up

information.

*$p < .05$ for the effect of time in mixed-effects model with random intercept

using Proc GLIMMIX; includes participants with missing data over time.